\pdfoutput=1

\documentclass[11pt]{article}

\usepackage[pagebackref=True]{hyperref}
\usepackage{EMNLP2023}

\usepackage{times}
\usepackage{latexsym}

\usepackage[T1]{fontenc}

\usepackage[utf8]{inputenc}

\usepackage{microtype}

\usepackage{inconsolata}

\usepackage[colorinlistoftodos]{todonotes}
\usepackage{algorithm}
\usepackage{algpseudocode}
\usepackage{multirow}
\usepackage{booktabs}
\usepackage{color}
\usepackage{pifont}
\usepackage{bbding}
\usepackage{afterpage}
\usepackage{cleveref}
\usepackage{xcolor}
\usepackage{textcomp}
\usepackage{listings}
\usepackage{xcolor}
\crefformat{section}{\S#2#1#3}
\crefformat{subsection}{\S#2#1#3}
\crefformat{subsubsection}{\S#2#1#3}
\newcommand{\ibold}[1]{\textbf{\textit{#1}}}

\newcommand{\checkmark}{\textcolor{green}{\Checkmark}}
\newcommand{\cross}{\textcolor{red}{\XSolid}}

\definecolor{pastelgreen}{RGB}{102,204,102}
\definecolor{pastelred}{RGB}{255,77,77}

\newcommand{\deltaup}[1]{\makebox[0pt][l]{\textcolor{pastelgreen}{\small\textuparrow#1}}}
\newcommand{\deltadown}[1]{\makebox[0pt][l]{\textcolor{pastelred}{\small\textdownarrow#1}}}

\algtext*{EndWhile} 
\algtext*{EndIf} 
\algtext*{EndFor} 

\title{MAF: Multi-Aspect Feedback for Improving \\ Reasoning in Large Language Models}

\author{Deepak Nathani , David Wang , Liangming Pan , William Yang Wang \\ 
University of California, Santa Barbara, Santa Barbara, CA \\
\texttt{\{dnathani, d\_wang, liangmingpan, william\}@cs.ucsb.edu}}

\begin{document}
\maketitle
\begin{abstract}

Language Models (LMs) have shown impressive performance in various natural language tasks. However, when it comes to natural language reasoning, LMs still face challenges such as hallucination, generating incorrect intermediate reasoning steps, and making mathematical errors. Recent research has focused on enhancing LMs through \textit{self-improvement} using feedback. Nevertheless, existing approaches relying on a single generic feedback source fail to address the diverse error types found in LM-generated reasoning chains. In this work, we propose \ibold{Multi-Aspect Feedback}, an iterative refinement framework that integrates multiple feedback modules, including frozen LMs and external tools, each focusing on a specific error category. Our experimental results demonstrate the efficacy of our approach to addressing several errors in the LM-generated reasoning chain and thus improving the overall performance of an LM in several reasoning tasks. We see a relative improvement of up to 20\% in Mathematical Reasoning and up to 18\% in Logical Entailment. We release our source code, prompts, and data\footnote{Our source code can be found \href{https://github.com/deepakn97/MAF/tree/main}{here}} to accelerate future research.

\end{abstract}

\section{Introduction}
\label{sec:intro}

Recent research in Language Models has focused on augmenting them with external tools \citep{schick2023toolformer, paranjape2023art}, learning from feedback \citep{Ouyang2022RLHFOpenAI, Akyurek2023RL4F} and iterative-refinement \citep{madaan2023selfrefine, Paul2023REFINER, Shinn2023ReflexionLA}. Iterative-Refinement has been a necessary tool in human evolution and problem-solving. Moreover, humans seek feedback from \textit{domain-specific} knowledge sources. For instance, an architect tasked with creating an environmentally friendly and structurally sound building design will require targeted feedback from civil engineers for structural integrity, and sustainability experts for eco-friendly design principles. However, previous works on iterative refinement overlook this requirement and collect \textit{generic} feedback from multiple sources or collect a single one from the Language Model itself.

\begin{figure*}[ht]
    \includegraphics[width=\textwidth]{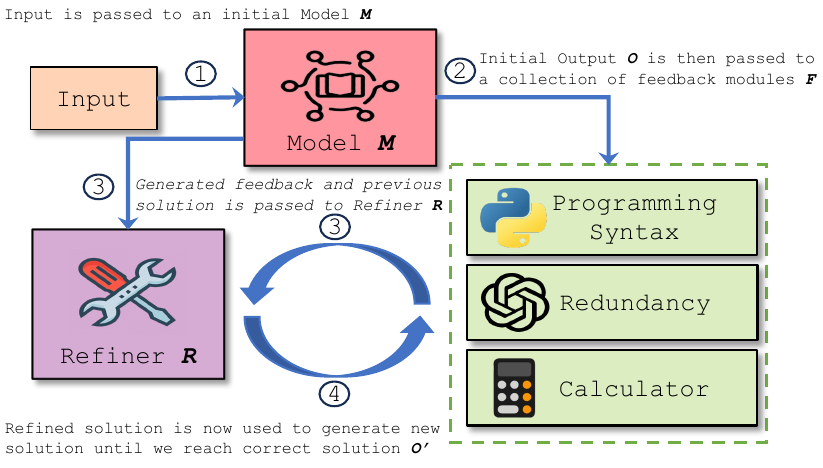}
    \centering\large
    \caption{Overview of the Multi-Aspect Feedback framework. We loop over steps 3 and 4 until we reach a final solution \ibold{O\textsuperscript{'}} or up to a set number of iterations.
    }
    \label{fig:maf}
\end{figure*}

In this work, we first investigate whether the use of generic feedback is a bottleneck in addressing the diverse range of errors present in the reasoning chains generated by LMs. We posit that utilizing generic feedback for a broad spectrum of errors may result in vague or non-actionable feedback for specific errors. This is due to two main factors: (i) the inclusion of multiple error categories within a single prompt, which not only increases the size of the prompt but also poses a challenge to current models that have a limited context length and struggle with long texts, and (ii) the model is burdened with identifying a multitude of error types in a single instance, which degrades the quality of the generated feedback.

To surmount these challenges, we introduce Multi-Aspect Feedback (MAF), a novel general-purpose iterative refinement framework that employs a collection of specialized feedback modules, including pre-trained LMs and external tools, each tailored to address a specific error category. Our framework includes a base model to generate an initial solution, multiple feedback modules, and a refiner model to revise the solution using feedback. Contrary to previous works, our feedback modules can use one of two refinement strategies: \underline{\texttt{Eager Refinement}} and \underline{\texttt{Lazy Refinement}} \cref{sec:refinement}. When using the Eager-Refine mode, the solution is revised immediately before moving on to the next feedback, however, for Lazy Refinement, we first collect feedback from multiple feedback sources and then refine the solution using this collective feedback.

In devising MAF, we first classified errors in LM-generated reasoning chains based on \citet{golovneva2022roscoe} and also identified some new error categories. Subsequently, we decoupled our feedback modules such that each module focuses on a single error category. This strategy not only elevates performance but also allows one to leverage specialized tools tailored for distinct error types, such as utilizing a code interpreter for syntax errors in generated Python programs instead of relying on generic LM feedback. Through a modular design, we ensure that each module of our framework is replaceable depending on the task. We conduct extensive experimentation on a variety of complex reasoning tasks such as Mathematical Reasoning \citep{shi2023gsmic}, Logical Reasoning \citep{2021entailmentbank}, and Multi-Hop Question Answering \citep{2019drop}, and establish that MAF effectively addresses the challenges previously discussed, and significantly enhances the performance of Language Models in complex reasoning tasks. We compare our method against the Base LMs and Self-Refine \citep{madaan2023selfrefine}, which also implements iterative refinement. During our experiments, we find that \texttt{Over-Refining} \cref{sec:experiments} the solution can lead to worse performance, and thus we treat the number of iterations as a hyperparameter. We also propose an Oracle Verifier setting \cref{sec:experiments}, where we assume we have access to an "answer verifier" and stop the refining process once we have reached the correct answer. To avoid unfair comparisons, we test Self-Refine under this setting and find that MAF provides actionable and specific feedback.

To summarize, the main contributions of this work are as follows:
\begin{enumerate}
\item We propose a novel framework that decouples the feedback generation process for different error categories. This allows us to use error-specific tools and LMs. Moreover, the proposed framework is \underline{modular} and all the modules are \underline{Plug-and-Play}. This allows for better flexibility on new tasks and the usage of advanced models as they become available.
\item Additionally, we propose two different types of refinement strategies, \texttt{Eager Refinement} and \texttt{Lazy Refinement}. Eager Refinement allows immediate revision of a solution by a feedback module to avoid conflicts, while Lazy Refinement improves efficiency by performing revisions from multiple feedback modules together. 
\item We show that our framework outperforms the base Language Models and similar Iterative Refinement baselines on several complex reasoning tasks such as Math reasoning (GSM-IC), Logical Reasoning (Entailment Bank), and Multi-hop Question Answering (DROP).
\end{enumerate}


\section{Related Work}
\label{sec:related}

\paragraph{Chain-of-Thought Reasoning}
There has been a plethora of research on prompting Large Language Models to improve their reasoning capabilities. \citet{wei2023chainofthought} found that prompting the models to generate a reasoning chain in few-shot setting before solving a problem improves the performance by a huge margin. Further, \citet{kojimaZeroShot} found that in zero-shot setting prefixing the solution generation with \textit{Let's think step-by-step} has the same effect as generating intermediate reasoning chain and improves the performance. Following this work, \citet{Wang2022SelfConsistencyIC} proposed sampling multiple outputs from the model and selecting the final answer by majority voting. \citet{zhou2023leasttomost} also showed that decomposing the original question into smaller sub-problems can give a performance boost.  \citet{Madaan2022LanguageMO} and \citet{chen2022programofthought} showed that models trained on code can be used to generate a \textit{program-of-thought} similar to a \textit{chain-of-thought}, which enables the model to use a language interpreter for translating mathematical calculations into code. In this work, we use LMs' in-context learning abilities to implement the LM-based feedback and refiner modules.

\paragraph{Tool Augmented LLMs} However, even with these advanced prompting techniques, the LMs still fail for problems that require external knowledge and are still prone to problems such as hallucination and incorrect reasoning. To circumvent these issues, several recent works have introduced the use of external tools such as calculator, code interpreter, knowledge retrieval \citep{karpas2022mrkl, chen2022programofthought, schick2023toolformer}. \citet{Yang2023GPT4ToolsTL, Patil2023GorillaLL, Shen2023HuggingGPTSA} show that one can teach LLMs to use tools by generating API calls. \citet{Li2023APIBankAB} also released a large dataset to enable research in the field of augmented Language Models. \citet{Hao2023ToolkenGPTAF} propose to represent each tool as a token and learn tool-specific embeddings to increase the robustness of these language models for using external APIs. However, reasoning is an iterative task and it is difficult to find a plausible answer for several problems in one-shot even when these LLMs are augmented with tools. This has inspired some recent works to use iterative refinement frameworks. In our work, we use external tools for generating feedback for certain error categories such as Programming Syntax Errors, Calculator for mathematical equations etc.

\begin{table}[ht]
\centering\small
    \begin{tabular}{c|c|c|c}
        \toprule
        \textbf{Task} & \textbf{Feedback Type} & \textbf{Type} & \textbf{ER}\\
        \midrule
        & Programming Syntax & Interpreter & \checkmark \\
        & Variable Naming & OpenAI & \checkmark\\
        \texttt{Math} & Redundancy & OpenAI & \cross \\
        \texttt{Reasoning}& Commonsense & OpenAI & \cross \\
        & Missing Step & OpenAI & \cross \\
        \midrule
        \texttt{Logical}& Redundancy & OpenAI & \cross \\
        \texttt{Reasoning}& Repetition & OpenAI & \cross \\
        & Hallucination & OpenAI & \cross \\
        \midrule
        & Redundancy & OpenAI & \cross\\
        \texttt{Question}& Factuality & OpenAI & \cross\\
        \texttt{Answering}& Commonsense & OpenAI & \cross\\
        & Missing Step & OpenAI & \cross\\
        \bottomrule
    \end{tabular}
    \caption{Feedback modules used for each task and their types. OpenAI type modules use a LLM to provide feedback. Feedback Modules with \underline{\texttt{Eager-Refine (ER)}} enabled, refine the solution without waiting for feedback from the other feedback modules. (\cref{sec:refinement}).}
    \label{tab:feemo}
\end{table}

\paragraph{Learning from Feedback} \citet{Schick2022PEERAC} took the first step towards iteratively fixing the problems and introduced the idea of training multiple instances of the same model to assist in different stages of problem-solving. It was introduced as a collaborative editing framework. \citep{madaan2023selfrefine} proposed to use the same model for generating initial solution, feedback generation, and refiner. Furthermore, recently there have been multiple works exploring the use of natural language feedback to improve performance. \citet{Shinn2023ReflexionLA} proposes converting different types of feedback into natural language and storing the feedback in a memory buffer to revisit later. \citet{Akyurek2023RL4F} train a critique generator to maximize the end-task performance of a larger base model using reinforcement learning. \citet{Paul2023REFINER} proposes a framework to fine-tune LMs for generating intermediate reasoning steps while interacting with a trained critique model to bridge the gap between small and large models. Recent research \citep{Wu2023FineGrainedHF} has also demonstrated the benefits of incorporating decoupled fine-grained feedback for each error category by using a set of fine-tuned reward models.

Despite these advancements, there is a lack of empirical evidence supporting the effectiveness of decoupled multi-aspect feedback for \textit{iterative refinement}. Our work addresses this gap by demonstrating that a generic feedback module, is insufficient to address the diverse range of potential errors in language model responses. To overcome this limitation, we propose a suite of feedback modules, each specifically targeting a particular error category and providing detailed feedback to improve both reasoning and solution quality.

\section{MAF: Multi-Aspect Feedback}
\label{sec:method}

\begin{figure*}[ht]
    \centering\large
    \includegraphics[width=\textwidth]{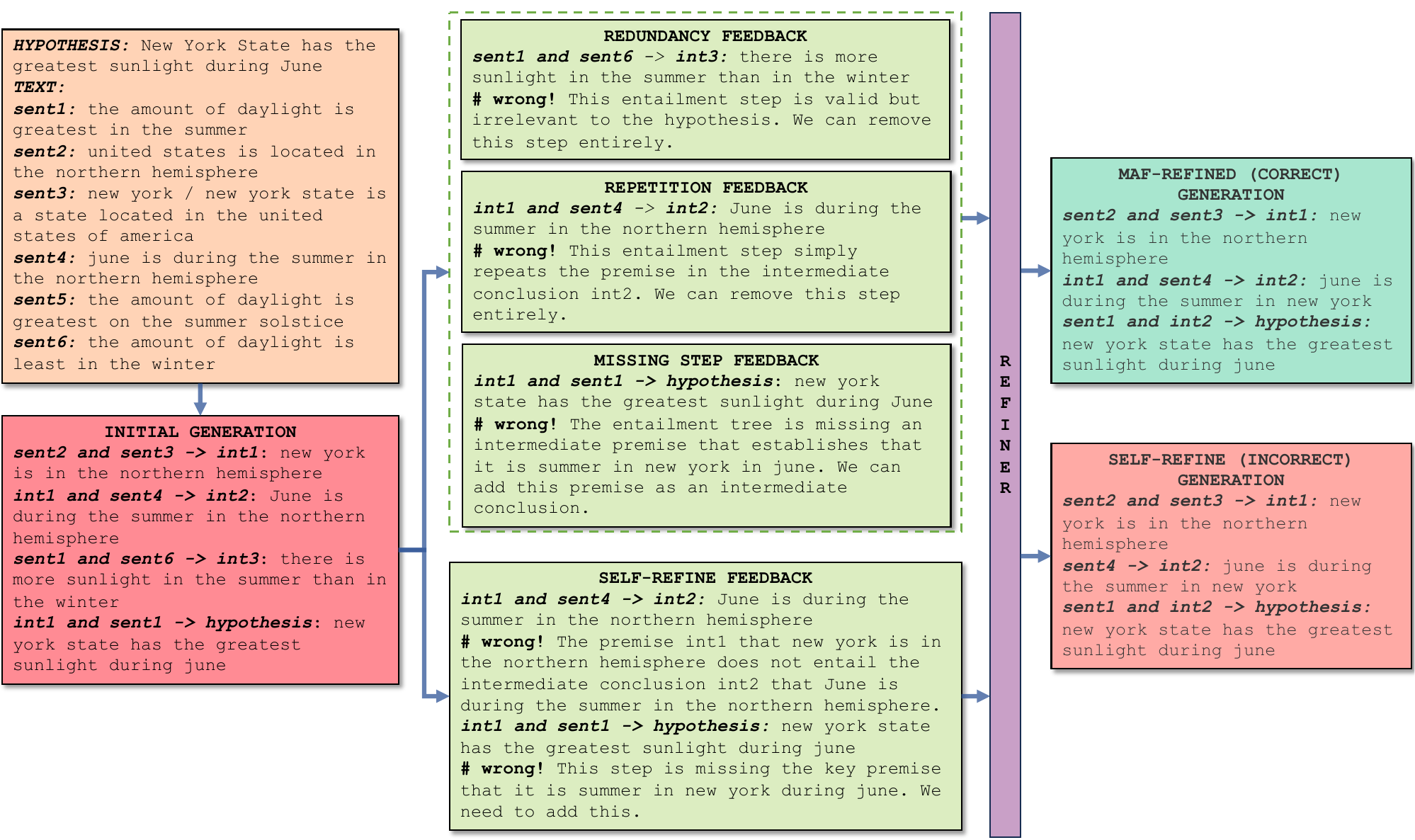}
    \caption{Comparison of the feedback generated by MAF and Self-Refine.}
    \label{fig:example}
\end{figure*}

\begin{algorithm*}
\caption{MAF algorithm}
\label{alg:maf}
\begin{algorithmic}[1]
\Require Input \textit{x}, model \ibold{M}, Refiner \ibold{R}, number of iterations $T$
\Require $n$ Eager-refine $\{s_1, s_2, ... s_n\}$ and $m$ Lazy-Refine $\{p_{1}, p_{2}, ... p_m\}$ Feedback Modules
\State Initialize output $y_{0}$ from \ibold{M}
\While{$i < T$}
    \For{$j \gets 1$ to $n$} \Comment{Eager-refine Feedbacks}
        \State{$fs_j \gets generate\_feedback(s_j, y_{i})$} \Comment{Generate Feedback}
        \If{$fs_j$ indicates revision is required}
            \State $y_{i} \gets revise(\ibold{R}, fs_j, y_i)$ \Comment{Revise solution}
        \EndIf
    \EndFor
    \State{$\ibold{F} \gets ""$} \Comment{Initialize empty feedback}
    \For{$k \gets 1$ to $m$} \Comment{Lazy-refine Feedbacks}
        \State{$fp_k \gets generate\_feedback(p_k, y_{i})$}
        \If{$fp_j$ indicates revision is required}
            \State{$\ibold{F} \gets \ibold{F} + fp_k$}
        \EndIf
    \EndFor
    \If{$\ibold{F}$ not empty}
        \State{$y_{i+1} \gets revise(\ibold{R}, fp_k, y_i)$}
    \Else
        \State{$y_{i+1} = y_i$}
    \EndIf
\EndWhile
\end{algorithmic}
\end{algorithm*}

\subsection{Overview}
In this work, we present an iterative refinement framework with an explicit focus on decoupling feedback generation for different error categories. This approach allows us to systematically address different error types in the generated solutions. Our proposed framework has three crucial components: a base language model \textbf{\textit{M}} that generates an initial solution \ibold{O}. A collection of $\textit{n}$ feedback modules $\{f_0, f_1, f_2 ... f_n\}$, each focusing on a single error category, collectively these modules generate a multi-aspect feedback \ibold{F}. And a Refiner \ibold{R} that generates refined solution \ibold{O\textsuperscript{'}} based on initial solution \ibold{O} and feedback \ibold{F}.

While there are other works that also use iterative refinement \citep{madaan2023selfrefine, Akyurek2023RL4F, peng2023check}, to the best of our knowledge, \underline{we are the first to explore the effect of decoupling} \underline{different types of feedbacks}. Taking inspiration from ROSCOE \citep{golovneva2022roscoe}, we categorize feedback into ten distinct categories: arithmetic, programming syntax, variable naming, missing step, coherency, redundancy, repetition, hallucination, commonsense, and factuality. Definitions of these error categories are provided in \autoref{sec:errorcat}. Moreover, a feedback module can be a tool such as a code interpreter for \texttt{syntax feedback}, a calculator for \texttt{arithmetic feedback}, a knowledge graph for \texttt{factuality or commonsense feedback}, a Language Model, or even a fine-tuned model. It is important to note that the feedback generation process is not limited to these categories and can be extended to include other categories as well. The overall process is illustrated in \autoref{fig:maf} and Algorithm \ref{alg:maf}.

\subsection{Initial Generation}
We use a large model such as GPT3.5\footnote{\href{https://openai.com/blog/chatgpt}{https://openai.com/blog/chatgpt}}, GPT4~\citep{OpenAI2023GPT4TR}, to generate an initial solution. However, generating just one solution isn't ideal for the reasoning process since reasoning is an iterative process. The solution often needs to be refined over time with every iteration bringing us closer to the correct answer. We follow the same principle in this work, where we initially produce a solution and then proceed to refine it based on actionable feedback.

\subsection{Feedback Modules} 
\label{sec:feedback}
Feedback generation is an involved task, providing a comprehensive list of common issues encountered in model outputs. The feedback generation is accomplished through a variety of tools known as feedback modules, which may include external tools, frozen LLMs, fine-tuned models, and scorers. These modules are used to provide actionable feedback based on the initial solution.

Each type of feedback module is suited to address specific types of errors. For instance, an external tool like a code interpreter would be ideal for providing feedback on syntax errors in code, while a fine-tuned model could provide more nuanced feedback on issues such as redundancy or hallucination. This decoupled approach allows us to address errors in a more targeted manner, improving the overall quality of the refined solution. \autoref{tab:feemo} shows all the feedback modules used in our work.

Furthermore, it's worth noting that merely specifying the error categories in a single prompt doesn't yield satisfactory results. We see two main reasons for this, firstly because the model is tasked with focusing on multiple errors simultaneously and secondly, as the number of error categories increase, the context length increases as well which results in high-quality feedback for the first few error categories and a steep decline in quality for the rest. While some of the previous works \citep{Wu2023FineGrainedHF, Paul2023REFINER} have explored using fine-grained feedback and achieved promising results.

Moreover, our feedback modules can choose between two refinement strategies: \texttt{Eager-Refine} or \texttt{Lazy-Refine}. We discuss this distinction in more detail in the next section. 

\subsection{Refinement} 
\label{sec:refinement}
In this work, we reuse the initial model \ibold{M} as a Refiner. During the refining phase, the refiner is given a solution and multi-aspect feedback and asked to revise the solution according to feedback. We found that these large models are proficient at the refining task, demonstrating a marked improvement in the quality of the final solution compared to the initial output. 

As mentioned in the previous section, we use two refinement strategies: \textit{Eager-Refinement} and \textit{Lazy-Refinement}. The eager-Refinement approach is used for feedback types that can cause conflicts during refinement. An example of such as feedback module would be \textit{Variable Naming} (\texttt{VN}), which is used to correct variable names in the generated code. This module can cause conflicts with others because when the other modules are referencing a variable that is supposed to be changed according to the \texttt{VN} feedback and can render the program inexecutable if refined incorrectly. Whereas, in the lazy refinement strategy feedback from the multiple modules is concatenated together along with the appropriate error categories to make a single multi-aspect feedback. This collective feedback is then passed to the Refiner model in order to get a revised solution. Another advantage of this approach is the increased efficiency by refining once for multiple errors and the added flexibility.

However, we also found that the smaller open-source models like LLaMA \citep{Touvron2023LLaMAOA}, Alpaca \citep{alpaca} and Vicuna\footnote{\href{https://lmsys.org/blog/2023-03-30-vicuna/}{https://lmsys.org/blog/2023-03-30-vicuna/}} often fail to adhere to the feedback provided thus making the refinement process ineffective. These smaller models also have smaller context lengths which make it difficult to include all the feedback in the prompt. To address this issue, we use a \texttt{Selective Summarization} approach. We only select parts of feedback that point to a problem. Figure \ref{fig:example} shows the "summarized" feedback generated by our approach. This simple selective summarization approach makes the feedback succinct and also proves to be less distractive to the models during the refining phase. This approach allows us to effectively combine all feedback together and use models with smaller context lengths to some extent.

\begin{figure*}[t]
    \begin{minipage}[c]{0.48\textwidth}
        \begin{tabular}{l|ccccc}
            \toprule
            \multicolumn{1}{c}{\textbf{Model}} & \textbf{EB} & \textbf{GSMIC} & \textbf{GSM8K} & \textbf{DROP} \\
            \midrule
            \texttt{GPT3.5} & $\textbf{56.1}$ & $76.2$ & $69.2$ & $\textbf{72.3}$\\
            $+$SR & $53.5$ \deltadown{2.6} & $77.0$ \deltaup{0.8} & $69.2$ \deltadown{0.0} & $62.0 \deltadown{8.3}$\\
            $+$SR\textsuperscript{\ding{72}} & $54.5$ \deltadown{1.6} & $87.0$ \deltaup{10.8} & $\underline{77.4}$ \deltaup{8.2} & $\underline{77.5}$ \deltaup{4.2}\\
            $+$MAF & $54.5$ \deltadown{1.6} & $\textbf{77.4}$ \deltaup{1.2} & $69.8$ \deltaup{0.6} & ${66.2}$ \deltadown{6.1} \\
            $+$MAF\textsuperscript{\ding{72}}  & $\underline{60.4}$ \deltaup{4.3} & $\underline{91.4}$ \deltaup{15.2} & $73.4$ \deltaup{4.2} & $76.4$ \deltaup{4.1}\\
            \midrule
            \texttt{ChatGPT} & $60.4$ & $72.0$ & $71.8$ & $\textbf{70.7}$\\
            $+$SR & $65.8$ \deltaup{5.4} & $76.0$\deltaup{4.0} & $\textbf{74.6}$ \deltaup{2.8} & $45.5$\deltadown{25.2}\\
            $+$SR\textsuperscript{\ding{72}} & $67.4$\deltaup{7.0} & $78.0$\deltaup{6.0} & $\underline{79.4}$ \deltaup{7.6} & $\underline{73.2}$\deltaup{2.5}\\
            $+$MAF & $\textbf{68.4}$\deltaup{8.0} & $\textbf{77.8}$\deltaup{5.8} & $73.2$ \deltaup{1.4} & ${67.9}$\deltadown{2.8}\\
            $+$MAF\textsuperscript{\ding{72}} & $\underline{71.7}$ \deltaup{11.3} & $\underline{82.8}$\deltaup{10.8} & $76.6$ \deltaup{4.8} & $72.7$\deltaup{2.0}\\
            \bottomrule
        \end{tabular}
        \captionof{table}{Experimental results for Entailment Bank (EB), GSMIC, GSM8K, DROP dataset as described in \cref{sec:experiments}. \textit{SR} represents Self-Refine \cite{madaan2023selfrefine}, and \textit{MAF} represents our method. \textsuperscript{\ding{72}} represents the Oracle Verifier setting (\cref{sec:results}). The best score for standard setting is in \textbf{bold} and \underline{underlined} for the Oracle Verifier setting.}
        \label{tab:results}
    \end{minipage}
    \hfill
    \begin{minipage}[c]{0.48\textwidth}
        \centering
        \includegraphics[width=\textwidth]{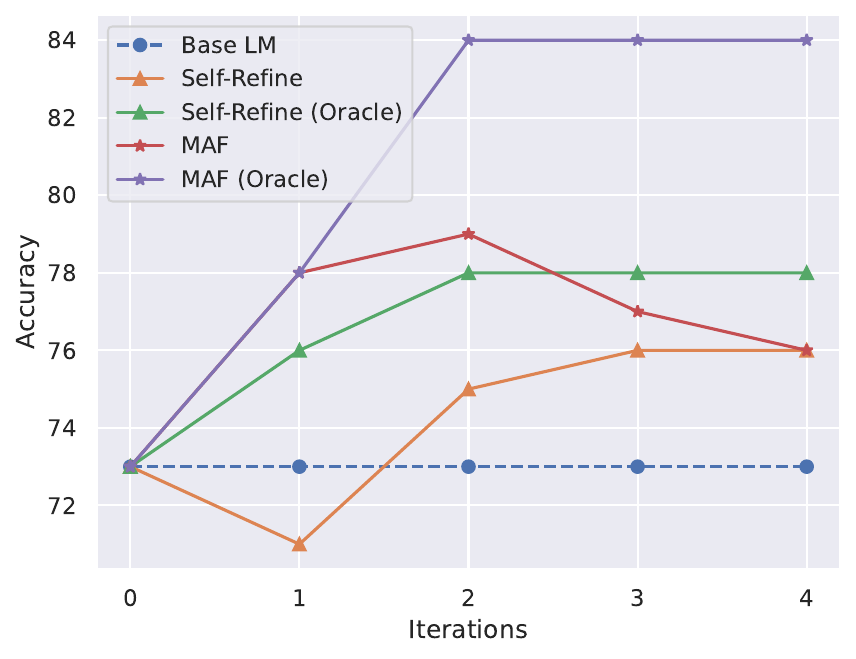}
        \caption{Accuracy for baselines and MAF on GSM-IC using ChatGPT under Standard and Oracle Verifier setting \cref{sec:results}.}
        \label{fig:overrefine}
    \end{minipage}
\end{figure*}





\section{Experiments} 
\label{sec:experiments}
\subsection{Datasets and Metrics}
\label{sec:datasets}
\paragraph{Mathematical Reasoning} The GSM8K dataset, presented by \citet{Cobbe2021GSM8K}, is a comprehensive compilation of high-quality grade school math problems. GSM8K has proven to be a great resource for testing LLMs on mathematical word problems, however, the performance on this dataset has been saturated for a while. 

To avoid that problem, we conduct experiments on a harder variant of this dataset, GSM-IC (Grade-School Math with Irrelevant Context), that was introduced by \citet{shi2023gsmic}. We run our main experiments on a randomly sampled subset of 500 problems from GSM-IC and use \% solve rate as the metric. 

\paragraph{Logical Reasoning} EntailmentBank, as described by \citet{2021entailmentbank}, is a dataset that contains multistep entailment trees. We use the validation set of Task 1 provided by the authors for our experimentation. For the metrics, we do not use the automated metrics provided by the original work because these metrics expect the trees to be similar in structure to gold trees. However, that is not a fair comparison because we find that there exist multiple correct entailment trees for a given hypothesis and information. Thus, we conduct a human evaluation and ask humans to evaluate if the hypothesis can be entailed from the predicted tree.

\paragraph{Question Answering} DROP \citep{2019drop} is a question-answering dataset. The Discrete Reasoning Over the Content of Paragraphs (DROP) dataset is designed for complex question-answering tasks that require multi-step reasoning over text passages. It presents a valuable benchmark for our iterative refinement framework, as the multi-step nature of the questions offers ample opportunities for generating feedback and refining to guide the model toward a correct solution. We use output parsing for answers and use \% correct answers as the final metric.

\subsection{Baselines}
\label{sec:baselines}

In this work, we focus on comparing our method with the Base LMs using the OpenAI API and a recently proposed iterative refinement framework, Self-Refine \citep{madaan2023selfrefine}. To provide a fair comparison and avoid randomness in the generated answer, \underline{we used Greedy Decoding for all our experiments}.

We follow \citep{madaan2023selfrefine} and use 8-shot Program of Thought \citep{chen2022programofthought} for GSM-IC since the Python program written by the model acts as a "calculator" for mathematical equations. 

The input for Entailment Bank includes a hypothesis and the supporting text, and we are limited by the context length of the current models. Hence, we use \underline{4-shot prompting} for this dataset. We use standard few-shot prompting \citep{Brown2020GPT3}, because the Entailment Tree itself acts like a reasoning chain.

Similarly, the examples in the DROP dataset have a passage and an accompanying question, so we use a \underline{3-shot Chain of Thought} \citet{wei2023chainofthought} prompting. We also provide an instruction specifying that the model should select either a number, date, or span from the passage to answer the question as shown in \autoref{sec:promptex}.

For Self-Refine, we use the prompts provided by the authors for GSM8K in their work for GSM-IC and write our own prompts for DROP and Entailment Bank. Self-Refine has three modules, initial generation, feedback, and refiner. We use the same parameters and prompting strategy as the corresponding baseline for the initial generation. The feedback module and refiner module are implemented using the standard few-shot prompting with 3 in-context examples for all datasets. Further details about the implementation of these baselines can be found in \autoref{sec:promptex}.

\subsection{Implementation}
\label{sec:implementation}

We implement MAF (Multi-Aspect Feedback) following the basic structure defined in \cref{sec:method} and Algorithm \ref{alg:maf}. The iterative refinement process continues until we reach the desired output quality or a task-specific stop criterion, up to a maximum of 4 iterations. Our method includes an initial generation, refiner, and feedback modules as shown in \autoref{tab:feemo}. The initial generation module uses the same parameters and prompting strategies as the corresponding baseline. Our Eager-Refine module uses 3-shot standard prompting and our Lazy-Refiner, which includes feedback from multiple modules, uses a 2-shot standard prompting approach. Our OpenAI-based feedback modules also use 3-shot standard prompting following \citep{madaan2023selfrefine} with an error-specific instruction.

To provide a fair comparison, we use the same prompts for Baselines and Self-Refine wherever possible. We use Greedy Decoding for all our OpenAI-based\footnote{Our experiments with GPT3.5 (Text-Davinci-003) and ChatGPT (gpt-3.5-turbo) are conducted using OpenAI API between April-2023 and August-2023.} modules to avoid any randomness. As mentioned in \cref{sec:feedback}, we use a \textit{selective summarization} approach to fit all the lazy-refine feedbacks in our refiner module. We use a basic rule-based strategy to summarize feedback. Since our feedback modules are instructed to inspect each intermediate step, they also include the steps with no mistakes in the generated feedback. However, feedback on these lines is not useful as such because there is no change in those steps. Thus we look for the steps with feedback \texttt{"looks good"} and remove those steps. This simple approach helps us increase the efficiency of our method by being able to include multiple feedbacks in our lazy-refiner.

\subsection{Results}
\label{sec:results}
In this section, we discuss the performance of our method and compare it to the baseline. We also discuss an ablation experiment studying the contribution of each feedback module. The main results of this work are shown in Table \ref{tab:results}. We find that our method can outperforms the base LMs and Self-Refine on a diverse set of reasoning datasets. For MAF, we report the results after $2$ iterations, and following \citet{madaan2023selfrefine} report the Self-Refine results after $4$ iterations.

\paragraph{GPT3.5 vs ChatGPT} We found that GPT3.5 can generate better feedback if an error is present in the solution, however often points out spurious errors even when they are not present. On the other hand, ChatGPT is more conservative in its feedback generation and often fails to detect the error even in an erroneous solution. Because of this conflicting behavior from the two models, our method is able to achieve similar performance using both models, even though ChatGPT is generally considered to be a better model than GPT3.5.

\paragraph{Over-Refining} Due to the behavior described above, we face the problem of \texttt{Over-Refining}. This means that once we have reached an optimal solution for a reasoning problem, forcing the LM to refine it further deteriorates the performance and the quality of the reasoning chain. As shown in Table \ref{tab:results} and \autoref{fig:overrefine}, Self-Refine \citep{madaan2023selfrefine} also suffers from this problem in the DROP dataset. Moreover, all the other iterative refinement framework such as Self-Refine uses a stop condition and stop the refining process if the generated feedback does not point out any problem. We however cannot take advantage of this stop condition because of the interplay between multiple feedback modules. For example, let's say our Missing Step module stops at iteration $k$ due to no missing steps at that point in time since the other feedback modules continue to refine the solution and self-refinement is an imperfect process, it might introduce a Missing Step error in further iterations. 

Hence, we treat the number of iterations as a hyperparameter and find that $2$ iterations work best for our method.

\paragraph{Oracle Verifier} Since the main focus of this work is to improve the incorrect initial generations from the model, we also evaluate the models under an Oracle Verifier setting. In this setting, we assume access to an "Oracle Verifier" which can judge the final answer generated by the model. If the final answer is correct, we stop the refinement process for that test sample, otherwise, we let the model continue refining the solution. It is important to note however that the model is not privy to this verification, and hence can still stop further refinement by not generating actionable feedback in the next iteration. Under this setting, we report results for both Self-Refine and MAF after 4 iterations. Even under this modified setting, we see that our method can outperform Self-Refine because we generate diverse feedback and thus can improve more solutions in the GSM-IC dataset.

\subsection{Ablation}
\label{sec:ablation}
In this section, we perform two ablation studies. Firstly, we analyze the impact of each feedback module, identifying those with the most significant effect on performance. Secondly, we compare the outcomes of our two proposed refinement strategies: \texttt{Lazy} vs \texttt{Eager} Refinement. The ablation studies are conducted on the GSM-IC dataset. Due to resource constraints, we conduct the following ablation studies on a smaller random subset of size 100.

\begin{table}[h]
    \begin{tabular}{l|cc}
        \toprule
        \midrule
        \textbf{Model} & Standard & Oracle \\
        \midrule
        ChatGPT & 73.0 & $-$ \\
        $+$MAF & 79.0 & 84.0 \\
        $-$Variable Naming (\texttt{VN})& 79.0 \deltadown{0.0} & 83.0 \deltadown{1.0} \\
        $-$Redundancy (\texttt{Red}) & 70.0 \deltadown{9.0}& 80.0 \deltadown{4.0} \\
        $-$Commonsense (\texttt{Com}) & 73.0 \deltadown{6.0} & 80.0 \deltadown{4.0} \\
        $-$Missing Step (\texttt{MS}) & 79.0 \deltadown{0.0} & 84.0 \deltadown{0.0} \\
        \midrule
        MAF (\texttt{VN}, \texttt{Red}, \texttt{Com}) & 79.0 & 85.0 \\
        \bottomrule
    \end{tabular}
    \caption{This table shows the contribution of each feedback module in MAF for the GSM-IC dataset. $-$ symbol in front of a module denotes the accuracy of our method after removing that feedback module. MAF (\texttt{VN}, \texttt{Red}, \texttt{Com}) shows the accuracy of our method using only the best-performing feedback modules.}
    \label{tab:ablation1}
\end{table}

\paragraph{Contribution of different Feedback Modules} We test the contribution of each feedback module by removing that module and calculating the accuracy under standard and Oracle verifier settings. While the standard setting highlights the potential negative impacts of some of the modules, the oracle verifier setting highlights the \textit{absolute} contribution of each module. This highlights an important finding that using the error categories is paramount to gaining performance. Results for this ablation study can be found in \autoref{tab:ablation1}.

We also calculate the accuracy of our method when using the three best-performing modules as the only source of feedback. We found that this does recover the performance of our method. Even though Variable Naming and Missing Step modules do not affect MAFs performance by a huge margin, it still makes our method more robust to a possible distribution shift. Moreover, the Variable Naming module's main contribution is not increasing the performance, but rather to increase the readability of the code and not confuse users with unclear names.

Note that we did not remove the Programming Syntax checker module as we use Program of Thoughts which requires a Python interpreter.

\paragraph{Lazy vs Eager Refinement} 

\begin{table}[h]
    \begin{tabular*}{\columnwidth}{l|ccc}
        \toprule
        \midrule
        \textbf{Model} & Standard & Oracle  \\
        \midrule
        \texttt{GPT3.5} & $\textbf{71.0}$ & $71.0$ \\
        $+$MAF & $67.0$ \deltadown{4.0} & $\textbf{85.0}$ \deltaup{14.0} \\
        $+$Only Lazy-Refine & $66.0$ \deltadown{5.0} & $81.0$ \deltaup{10.0} \\
        $+$Only Eager-Refine & $66.0$ \deltadown{5.0} & $\textbf{85.0}$ \deltaup{14.0} \\
        \midrule
        \texttt{ChatGPT} & $73.0$ & $73.0$ \\
        $+$MAF & $\textbf{78.0}$ \deltaup{5.0} & $\textbf{81.0}$ \deltaup{8.0} \\
        $+$Only Lazy-Refine & $69.0$ \deltadown{4.0} & $80.0$ \deltaup{7.0} \\
        $+$Only Eager-Refine & $73.0$ \deltadown{0.0} & $80.0$ \deltaup{7.0} \\
        \bottomrule
    \end{tabular*}
    \captionof{table}{Accuracy on GSM-IC when using different refinement strategies under Standard and Oracle settings. \textit{MAF} shows the performance of our method when using both Lazy and Eager refine in tandom. \textit{Only Lazy-Refine} means all feedback modules use Lazy-Refine and similarly \textit{Only Eager-Refine} means all modules use Eager-Refine strategy. The best score for each setting is in \textbf{bold}.}
    \label{tab:ablation2}
\end{table}

To illustrate the complementary nature of our two proposed refinement strategies, we evaluate the performance of our iterative refinement framework using all feedback modules in either Lazy or Eager mode. The results are presented in \autoref{tab:ablation2}, which showcases the performance of our framework under different refinement settings. In this table, 'MAF' corresponds to the results obtained by combining both Lazy and Eager Refinement strategies, as defined in \autoref{tab:feemo}, while 'Only Lazy/Eager-Refine' displays the results of our framework when utilizing only one type of refinement strategy.

The results clearly demonstrate that our framework, which leverages a combination of eager and lazy refinement, consistently matches or outperforms using either strategy in isolation, across both the standard and oracle verifier settings.

Practical considerations also favor the use of a hybrid approach incorporating both eager and lazy feedback. Relying solely on lazy feedback can lead to a situation where multiple feedback categories are condensed into a single prompt. Despite our feedback summarization technique, this can result in the iteration prompt exceeding the context window limit of many widely available models. Conversely, exclusively employing eager feedback may result in rewriting the solution for each feedback module, leading to high token usage. While using all modules in Eager-Refine mode can closely approach the performance of 'MAF' (as shown in \autoref{tab:ablation2}), it is not scalable when dealing with a large number of feedback modules.

\section{Conclusion}

In this work, we present Multi-Aspect Feedback (MAF), a novel iterative refinement framework that decouples the feedback modules and takes advantage of error-specific tools to generate feedback. We demonstrate the performance of our framework on a set of diverse reasoning tasks and compare it with other iterative refinement baselines. 

Contrary to previous works, we found that \textit{Over-Refinement} can be a problem in iterative refinement frameworks since models are not certain if their own answer is correct. Our work also draws focus on the necessity to devise better feedback methods and call for augmenting Language Models with them. We hope this work will inspire further research in this area and to this end, we make all our code, data, and prompts available.

\section*{Limitations}
The main limitation of our approach is that the base models need to have sufficient in-context learning abilities to process the feedback and refine the solution. Even with in-context learning abilities, these models are not perfect and thus can still make mistakes while refining the solution even when correct feedback is given.

All the experiments conducted in this work use large powerful models provided by OpenAI. We find that open-source LMs such as Vicuna, and Alpaca can generate decent initial solutions but are not capable of refining their own solutions. Thus, we leave the investigation of improving open-source models to future work.

Another limitation inherent in our approach is the reliance on a fixed set of Feedback Modules. Our method necessitates the pre-selection of feedback modules before execution, which in turn demands human intervention to determine the appropriate feedback categories for each new dataset or domain. Future research could explore novel methods that can dynamically and autonomously determine the most suitable feedback modules for specific problems in real-time.

\section*{Ethics Statement}

The experiments in this work were performed with models that are not open-sourced and are ever-changing. Moreover, these models are expensive to use, and thus research using these models requires an enormous amount of funding. The existing literature lacks details about the datasets that are being used to train these huge models or the filtering mechanism that is being used to clean the polluted corpora.

Furthermore, there is always a possibility for bad actors to use our method to generate more toxic or harmful text. Our approach does not guard against this.

\section*{Acknowledgements}
This work was supported by the National Science Foundation award \#2048122. The views expressed are those of the authors and do not reflect the official policy or position of the US government. Additionally, we thank our reviewers for their detailed and useful feedback.

\newpage
\bibliographystyle{acl_natbib}
\bibliography{anthology,custom}

\appendix
\label{sec:appendix}
\section{Error Categories}
\label{sec:errorcat}
In this section we define all the error categories that are implemented in this work. \autoref{tab:errorcat} shows the error categories introduced by \citet{golovneva2022roscoe}.

\subsection{Programming Syntax Feedback}
Programming syntax feedback module is implemented as Python Interpreter. This module aims to fix any syntax errors in the generated code. This particular module benefits from a Eager-Refinement approach.

\subsection{Variable Naming Feedback}
Good variable names in a code can improve the readability of the code and potentially can improve model's own understanding of the program in further iterations. Variable Naming Feedback is another module which benefits from an eager-refine approach.

\subsection{Redundancy Feedback}
Redundant information is any information included in the reasoning that doesn't help answer the question. This additional information may distract the model from correctly answering and should thus be removed.

\subsection{Commonsense Feedback}
Commonsense reasoning errors are errors about any relation or knowledge that is should be known from general world such as "all ducks are birds".

\subsection{Missing Steps}
Missing steps errors are any gaps in reasoning or missing information that prevent the reasoning chain from being correct. This also identifies the model saying that the question is unanswerable as a missing step error because that means additional reasoning steps are needed to answer the question from the passage.

\subsection{Factuality Feedback}
Factuality errors are cases where the answer reasoning states infactual information. This could be information that contradicts information given in the passage or hallucinated.

\subsection{Hallucination Feedback}
LLMs are prone to hallucination, however, it has been shown that sampling from a LLM multiple times and then selecting the majority answer can improve the results. Thus Hallucination feedback aims to fix any hallucinated facts in the initial generation. This module can be improved by using an external tool such as a Knowledge Source instead of a LLM.

\section{Implementation Parameters}
\autoref{tab:srparam} provides parameters used for Self-Refine and \autoref{tab:mafparam} provides parameters used for MAF.

\section{Few-Shot Prompt Examples}
\label{sec:promptex}
We add samples for all the prompts used in our work. Complete prompts can be found in our source code. Note that for all feedback prompt examples, there is at least one example with no errors, in which case the feedback will state that there are no errors and the reasoning is correct. This helps decrease the likelihood that the model identifies an error in a solution that is actually correct.

\newpage
\begin{table*}[ht]
    \centering\small
    \begin{tabular*}{\textwidth}{@{\extracolsep{\fill}}p{0.3\textwidth}|p{0.65\textwidth}}
        \toprule
        \textbf{Error Type} & \textbf{Definition} \\
        \midrule
        \textbf{Programming Syntax} & Syntax errors in code\\
        \textbf{Arithmetic} & Error in math calculations\\
        \textbf{Grammar} & Faulty, unconventional, or controversial grammar usage\\
        \textbf{Coherency} & Steps contradict each other or do not follow a cohesive story\\
        \textbf{Variable Naming} & Variable names in a program don't give full information or are wrong\\
        \textbf{Repetition} & Step paraphrases information already mentioned in previous reasoning steps\\
        \textbf{Hallucination} & Information is not provided in the problem statement and is irrelevant or wrong\\
        \textbf{Commonsense} & Model lacks relations that should be known from the general world (e.g., "1 dozen = 12")\\
        \textbf{Factuality} & Information about an object (i.e. quantity, characteristics) or a named entity doesn’t match with the input context\\
        \textbf{Missing Step} & The content of the generated reasoning is incomplete and lacks the required information to produce the correct answer\\
        \textbf{Redundancy} & Explanation contains redundant information, which even though might be factual, is not required to answer the question\\
        \bottomrule
    \end{tabular*}
    \caption{Error types as defined in \citet{golovneva2022roscoe}. Each error category is defined for a single step in the reasoning chain.}
    \label{tab:errorcat}
\end{table*}

\begin{table*}[ht]
    \centering
    \begin{minipage}{0.45\textwidth}
    \centering
    \begin{tabular}{l|ccc} 
        \toprule
        \midrule
        \textbf{Dataset} & \textbf{Base LM} & \textbf{Feedback} & \textbf{Refiner} \\
        \midrule
        EB & 300 & 600 & 600 \\
        GSM-IC & 300  & 600 & 600   \\
        GSM8K & 300 & 600 & 600 \\
        DROP  & 450 & 600 & 800  \\
        \bottomrule
    \end{tabular}
    \caption{Value of Maximum Number of Tokens parameter for Self-Refine on various datasets.}
    \label{tab:srparam}
    \end{minipage}
    \hfill
    \begin{minipage}{0.45\textwidth}
    \begin{tabular}{l|ccc} 
        \toprule
        \midrule
        \textbf{Dataset} & \textbf{Base LM} & \textbf{Feedback} & \textbf{Refiner} \\
        \midrule
        EB & 300 & 600 & 600 \\
        GSM-IC & 300  & 600 & 600 \\
        GSM8K & 300 & 600 & 600 \\
        DROP  & 450 & 600 & 800  \\
        \bottomrule
    \end{tabular}
    \caption{Value of Maximum Number of Tokens parameter for MAF on various datasets.}
    \label{tab:mafparam}
    \end{minipage}
\end{table*}

\clearpage

\begin{figure*}
\centering\small
\includegraphics[width=\textwidth]{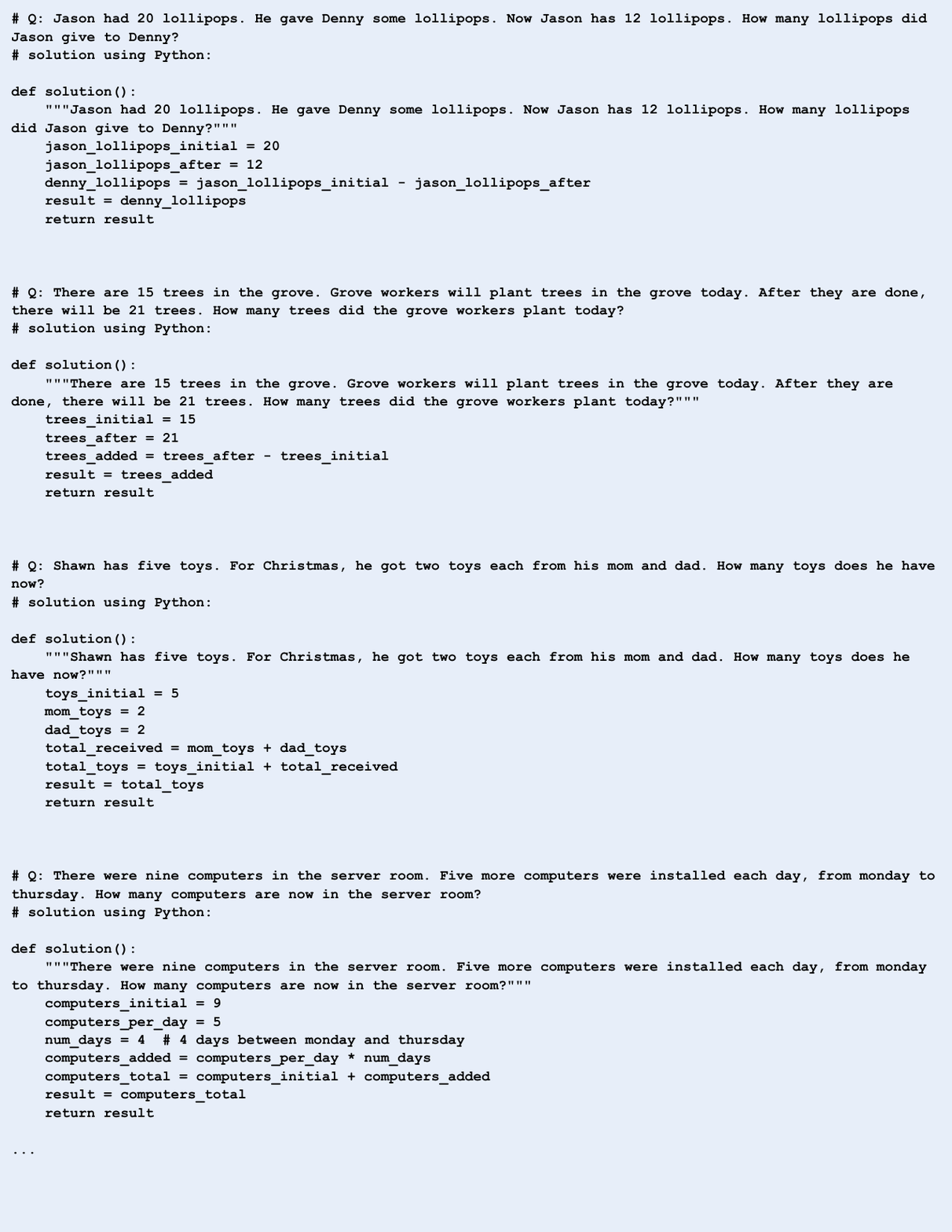}
\caption{Initialization prompt for Mathematical Reasoning}
\end{figure*}

\begin{figure*}
\centering\small
\includegraphics[width=\textwidth]{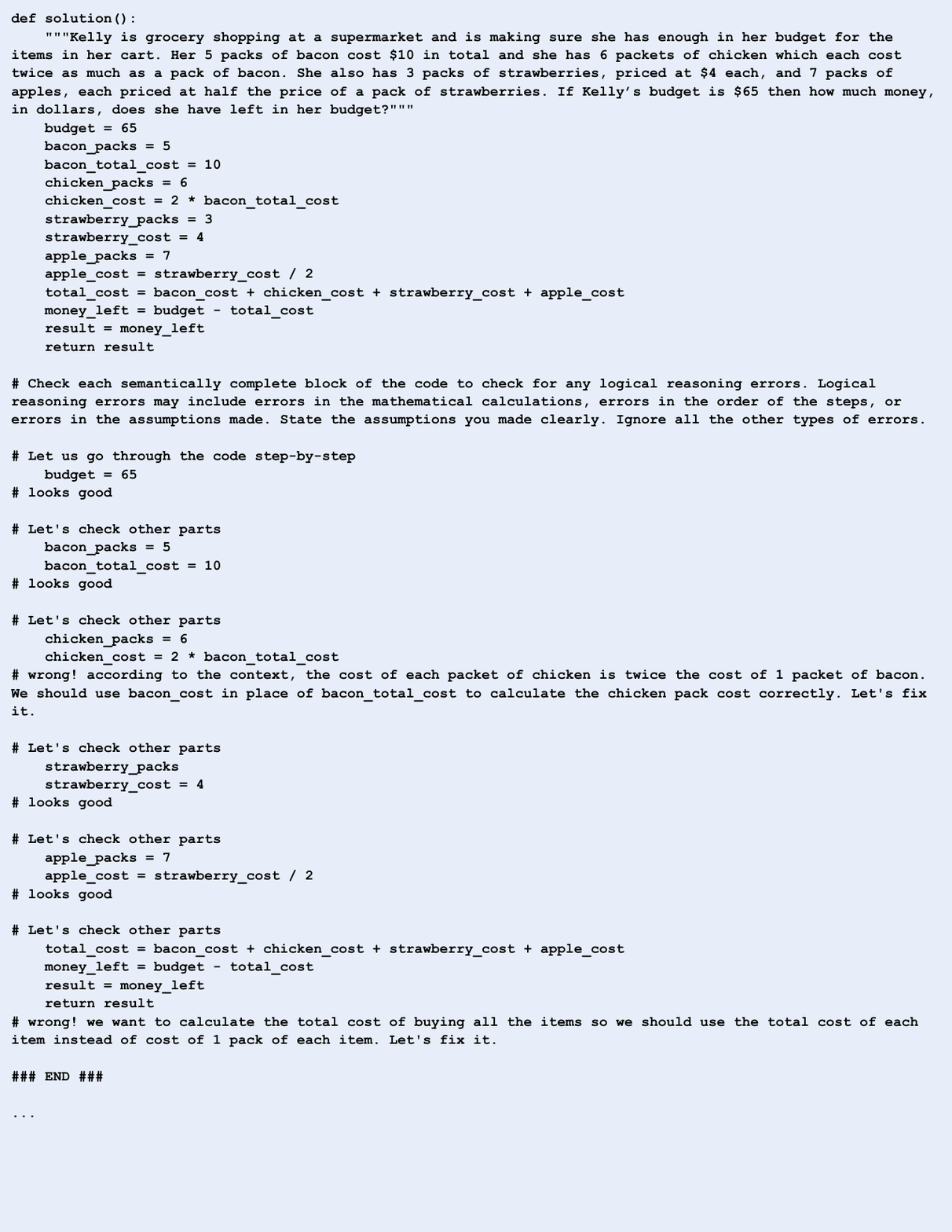}
\caption{Commonsense Feedback for Mathematical Reasoning}
\end{figure*}

\begin{figure*}
\centering\small
\includegraphics[width=\textwidth]{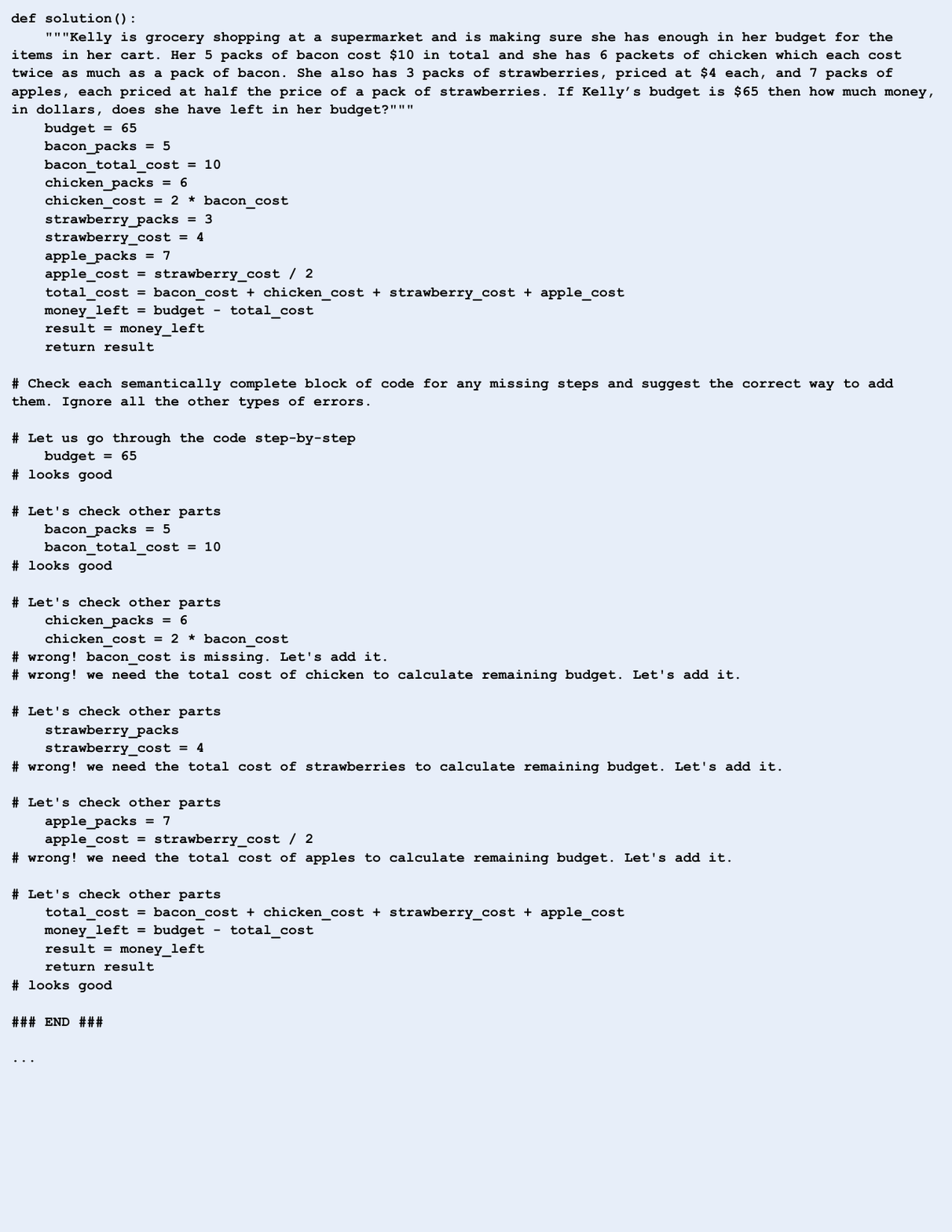}
\caption{Missing Step Feedback for Mathematical Reasoning}
\end{figure*}

\begin{figure*}
\centering\small
\includegraphics[width=\textwidth]{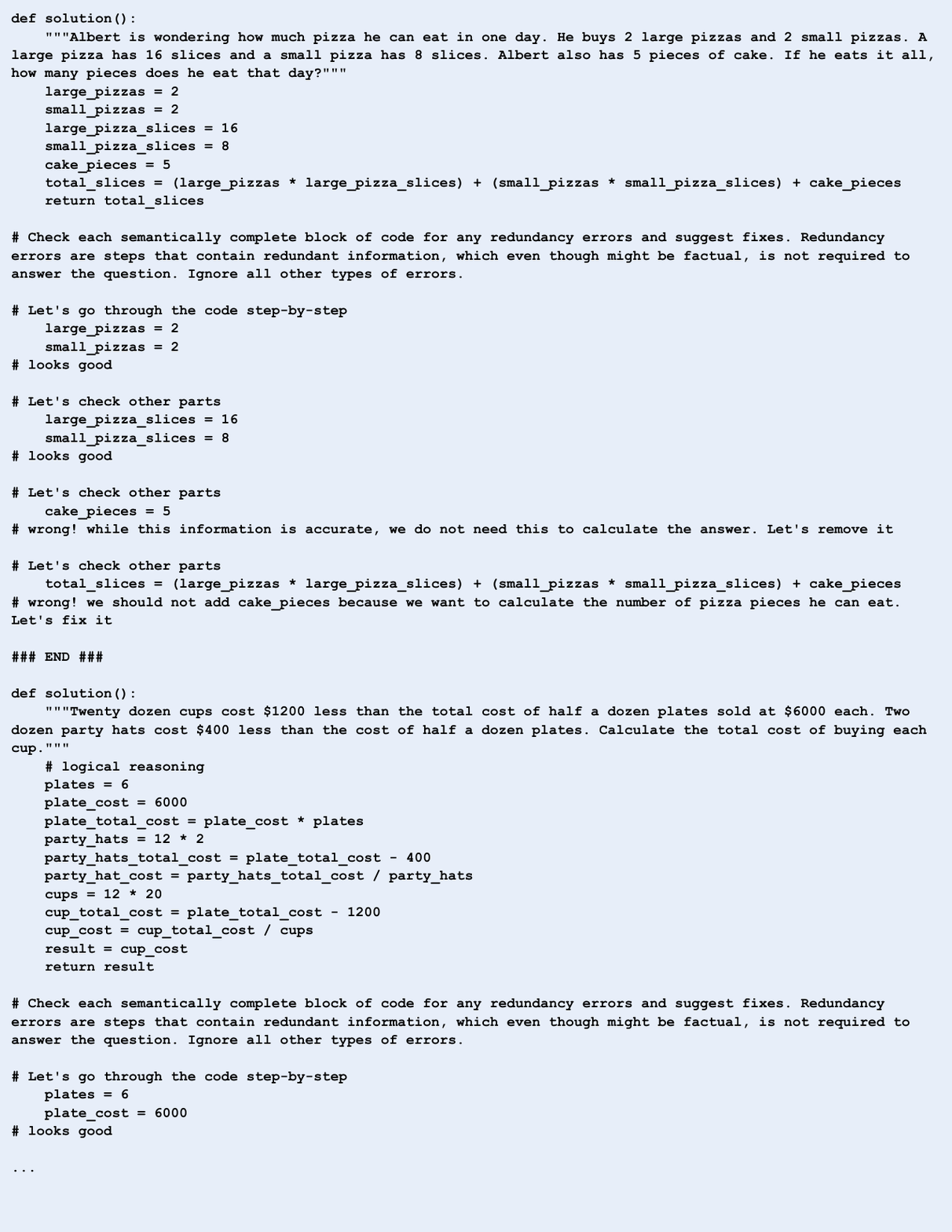}
\caption{Redundancy Feedback for Mathematical Reasoning}
\end{figure*}

\begin{figure*}
\centering\small
\includegraphics[width=\textwidth]{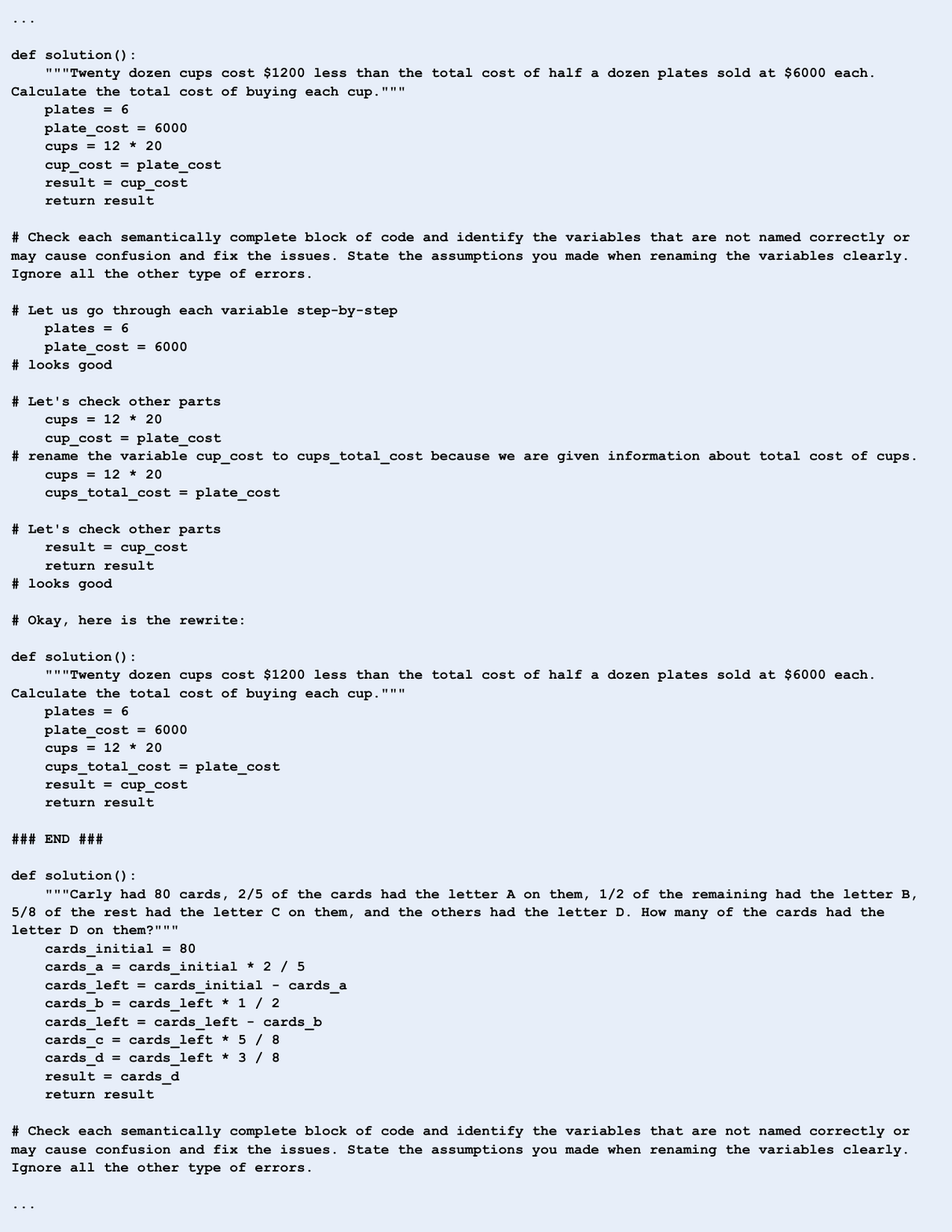}
\caption{Variable Naming Feedback for Mathematical Reasoning}
\end{figure*}

\begin{figure*}
\centering\small
\includegraphics[width=\textwidth]{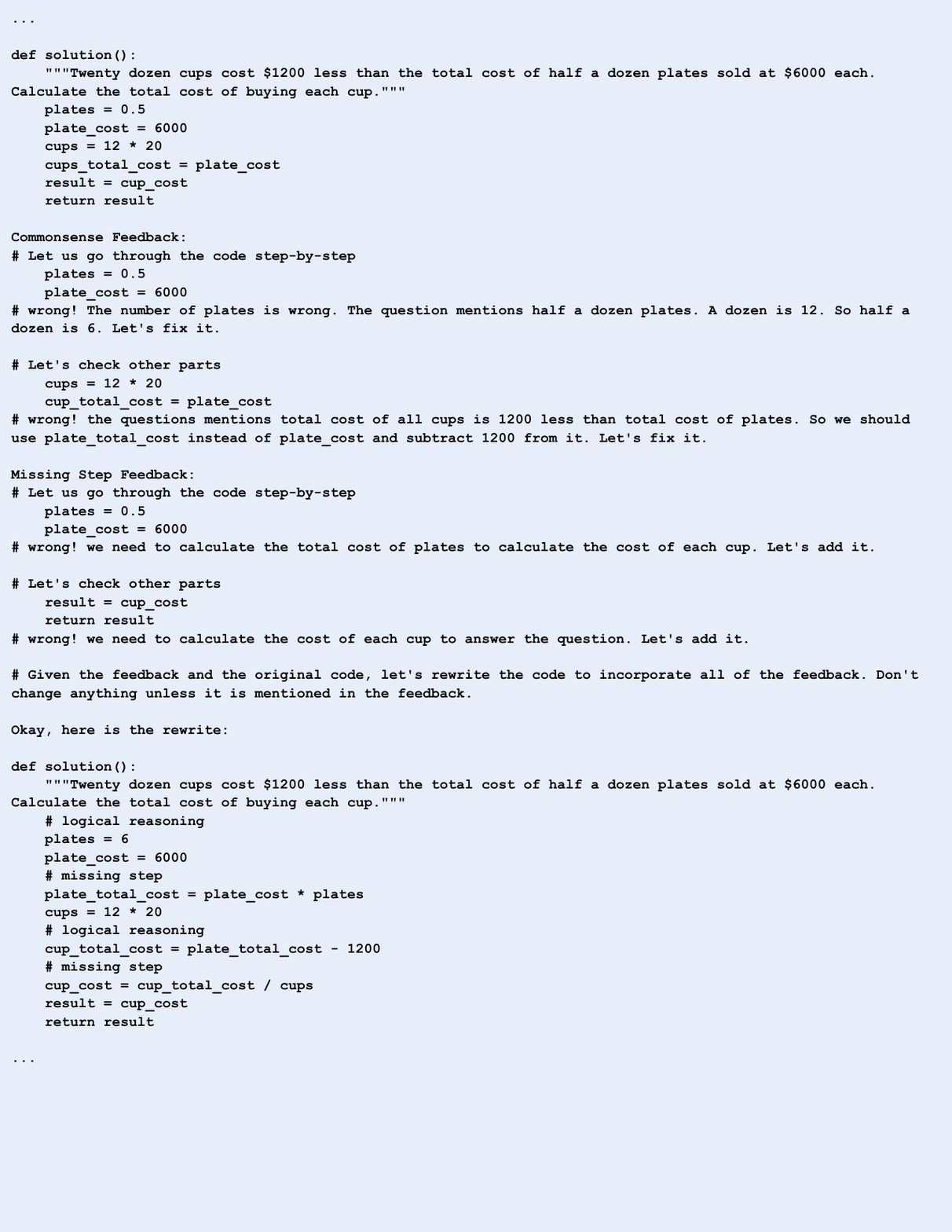}
\caption{Our Iterative Refinement prompt for Mathematical Reasoning}
\end{figure*}

\begin{figure*}
\centering\small
\includegraphics[width=\textwidth]{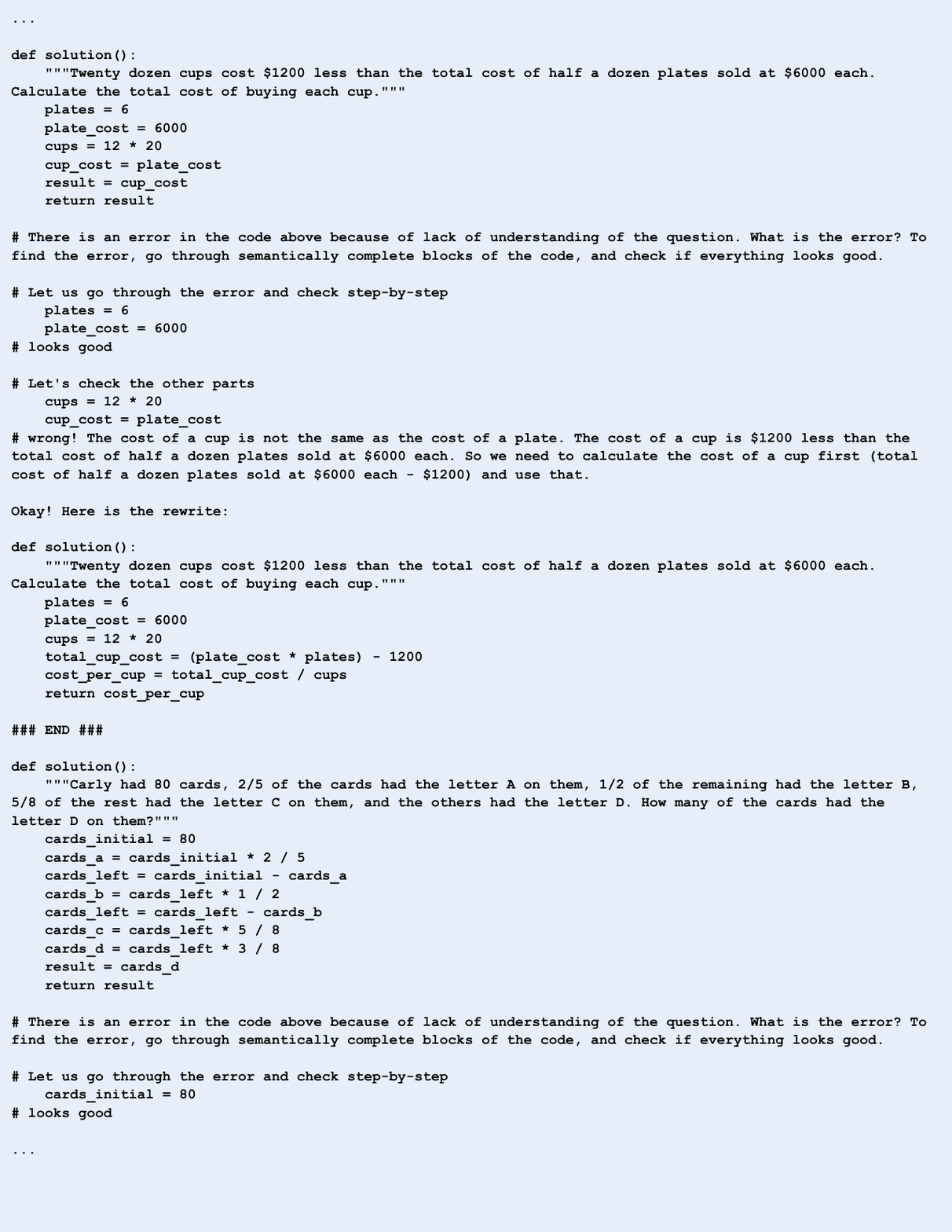}
\caption{Iterative Refinement prompt from Self-Refine for Mathematical Reasoning}
\end{figure*}

\begin{figure*}[htp]
\centering
\includegraphics[width=\textwidth]{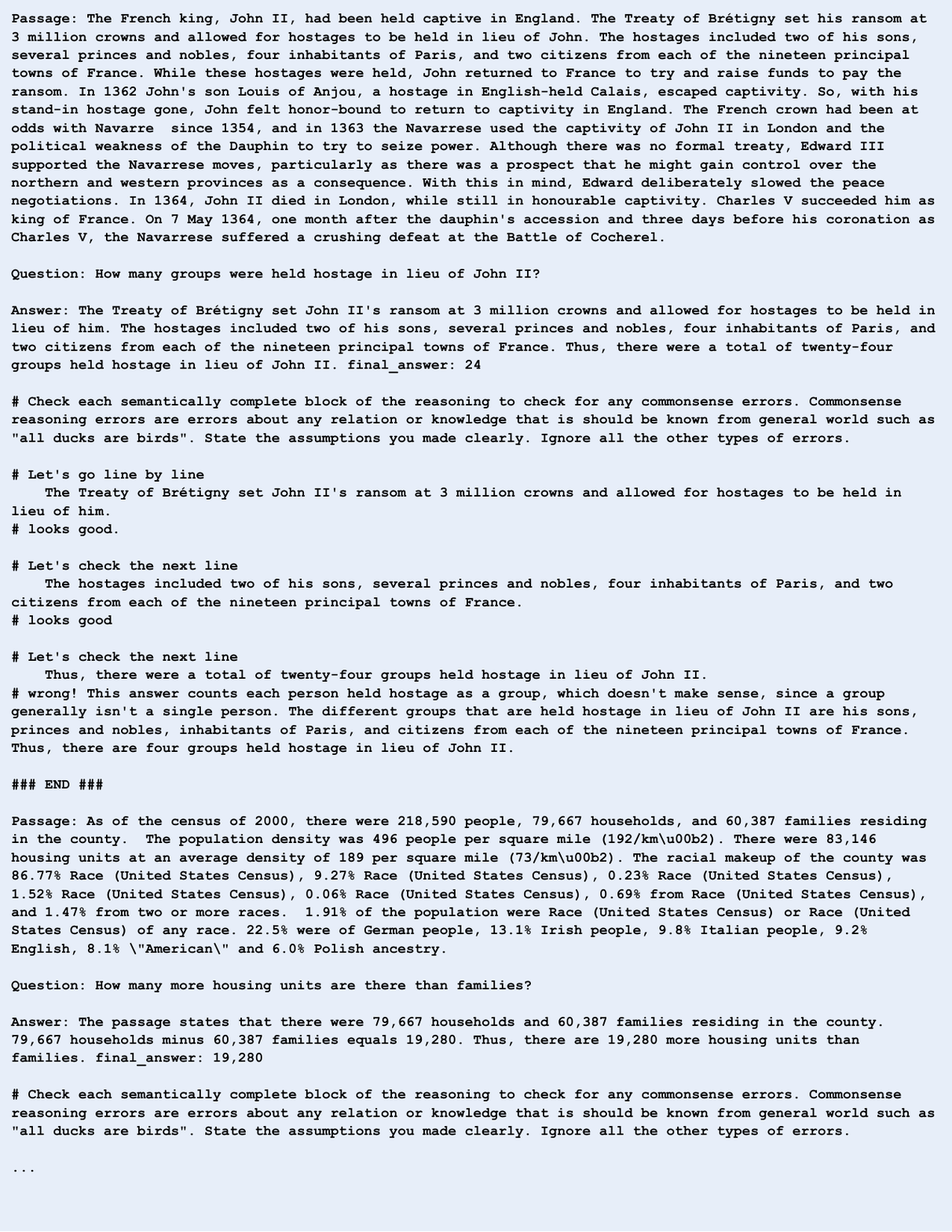}
\caption{Commonsense Feedback for Question Answering}
\end{figure*}

\begin{figure*}
\centering\small
\includegraphics[width=\textwidth]{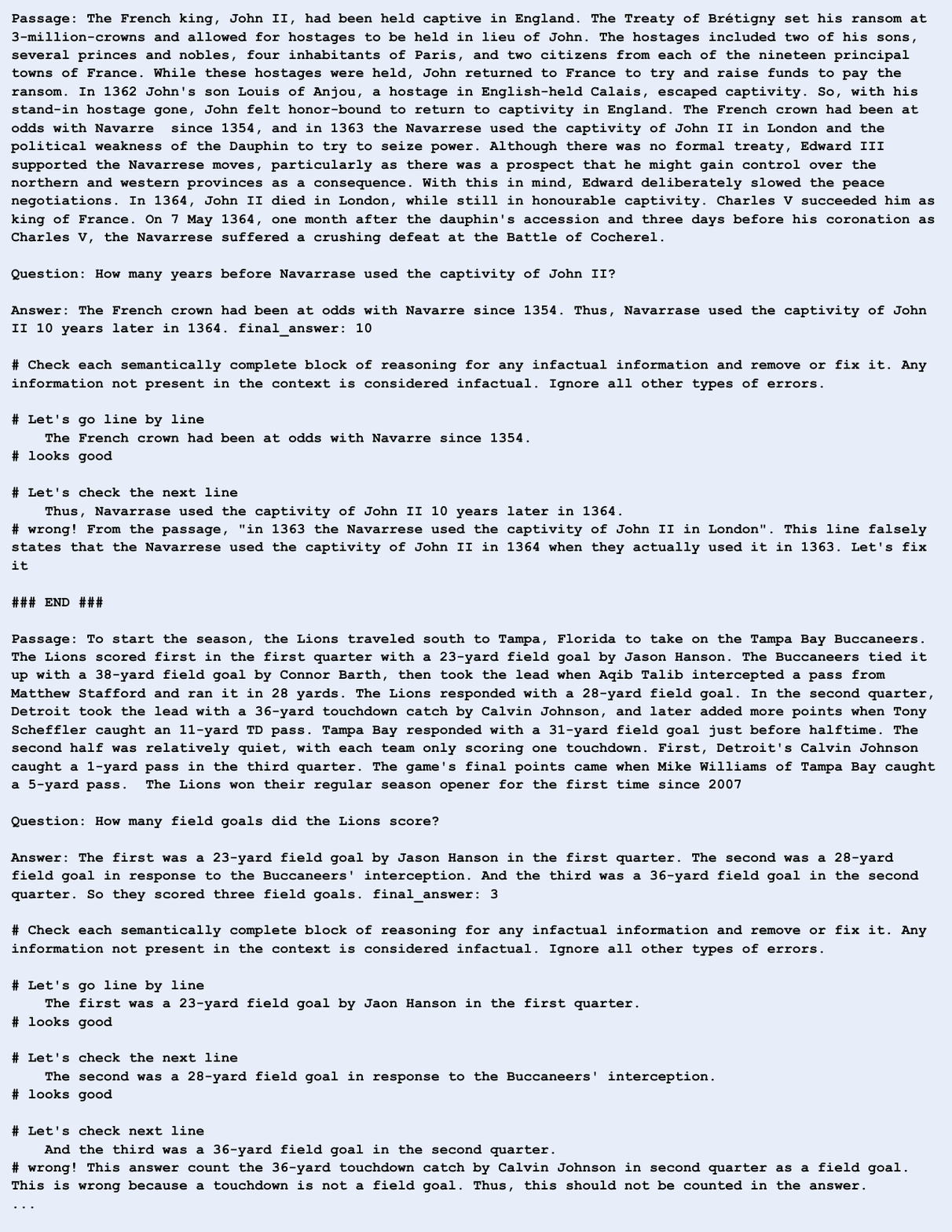}
\caption{Factuality Feedback for Question Answering}
\end{figure*}

\begin{figure*}
\centering\small
\includegraphics[width=\textwidth]{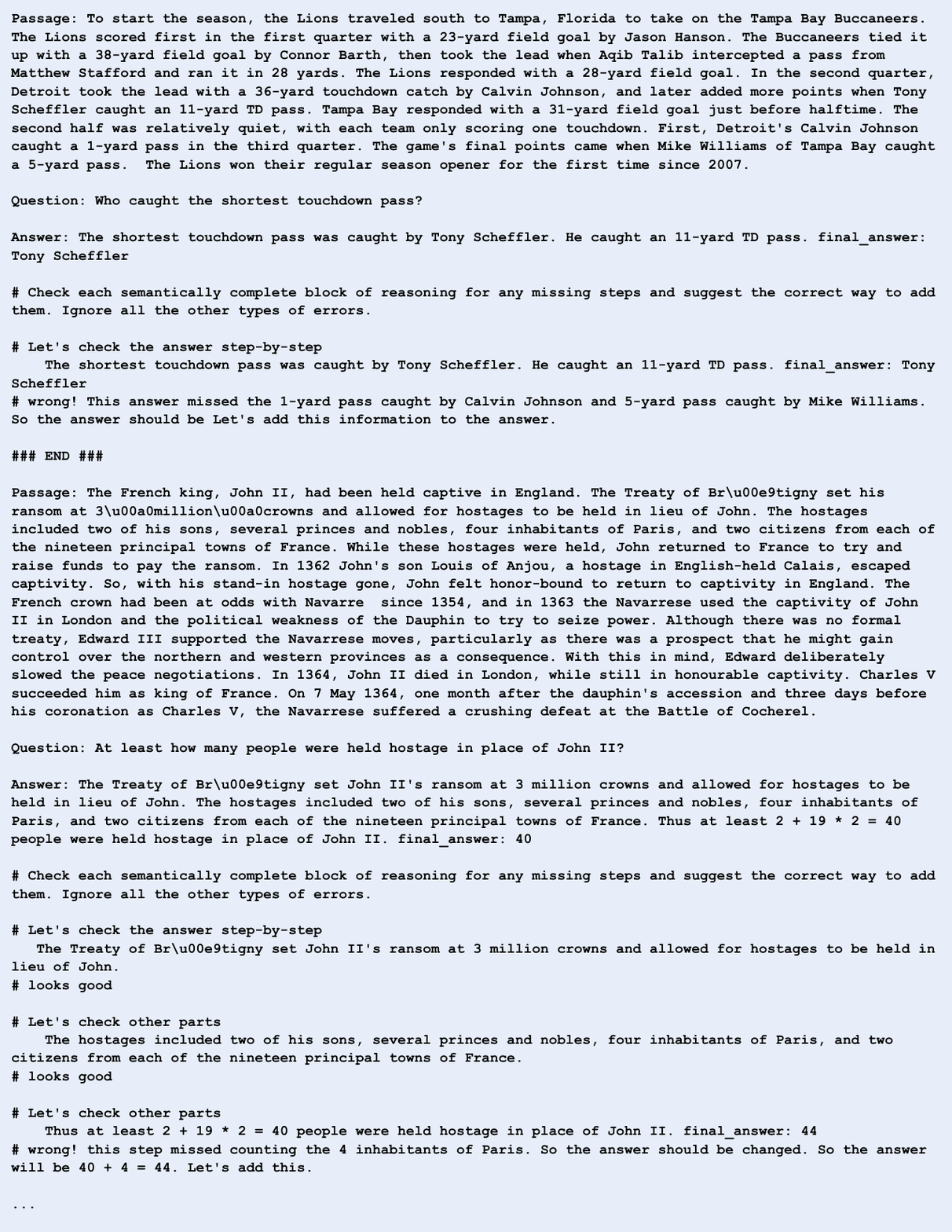}
\caption{Missing Step Feedback for Question Answering}
\end{figure*}

\begin{figure*}
\centering\small
\includegraphics[width=\textwidth]{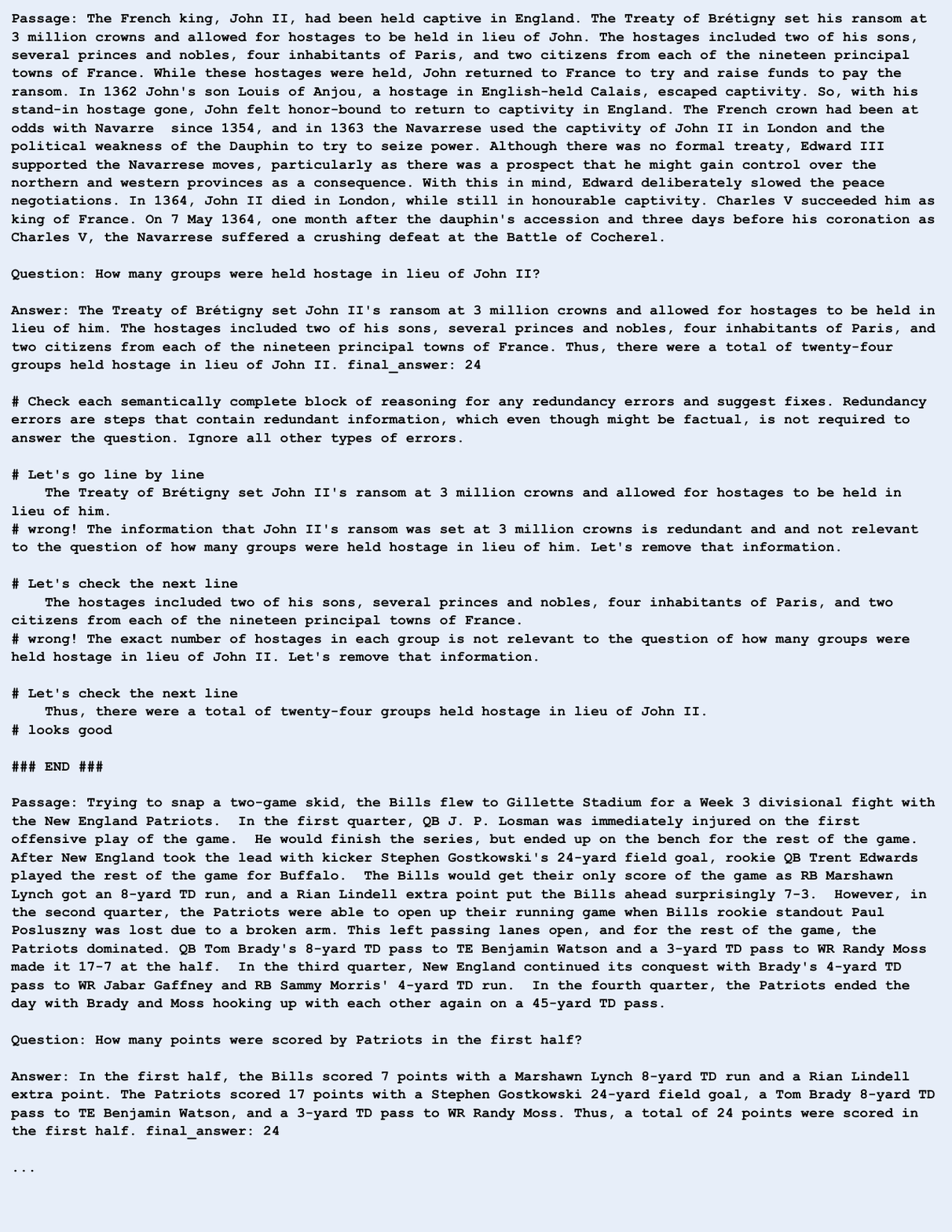}
\caption{Redundancy Feedback for Question Answering}
\end{figure*}

\begin{figure*}
\centering\small
\includegraphics[width=\textwidth]{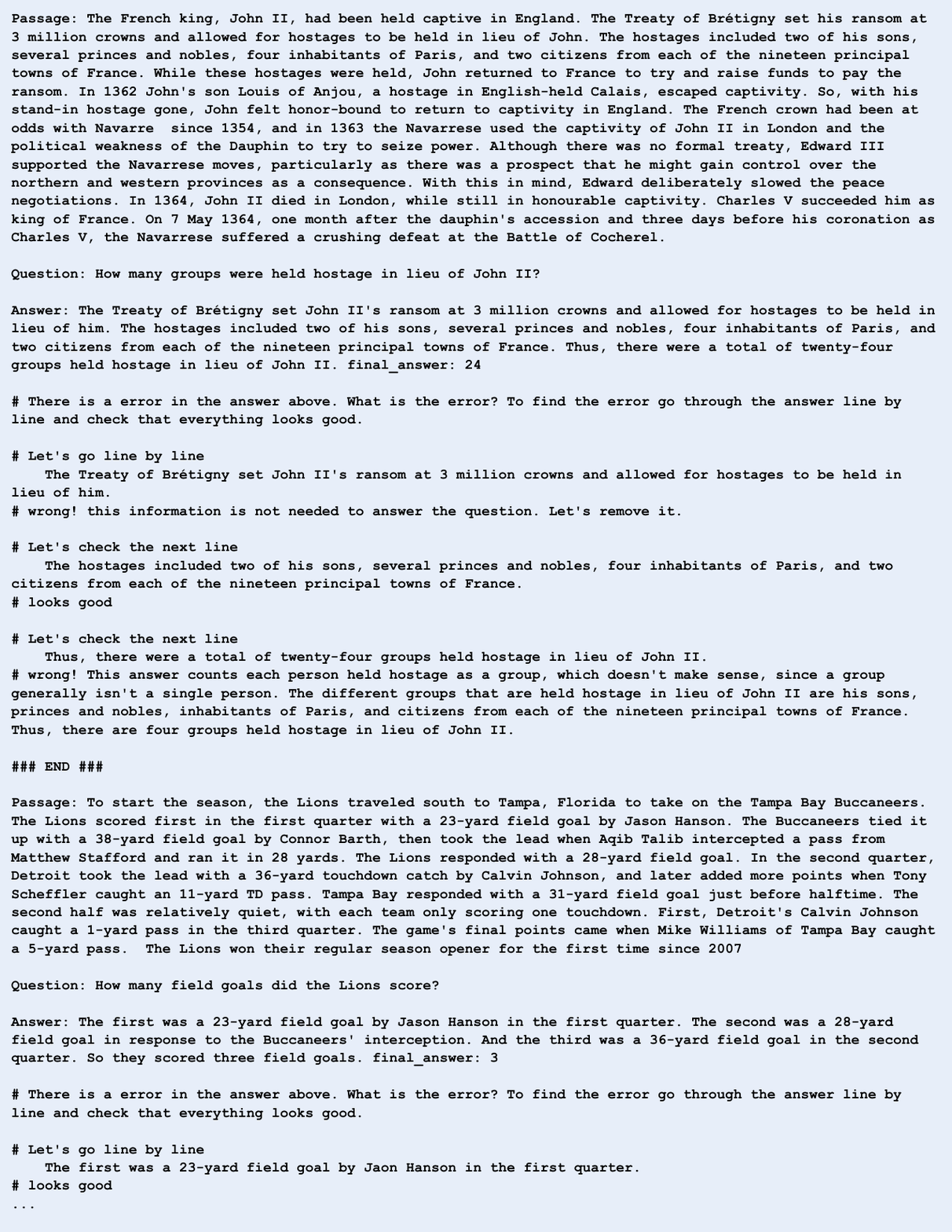}
\caption{Self-Refine style Iterative Refinement prompt for Question Answering}
\end{figure*}

\begin{figure*}
\centering\small
\includegraphics[width=\textwidth]{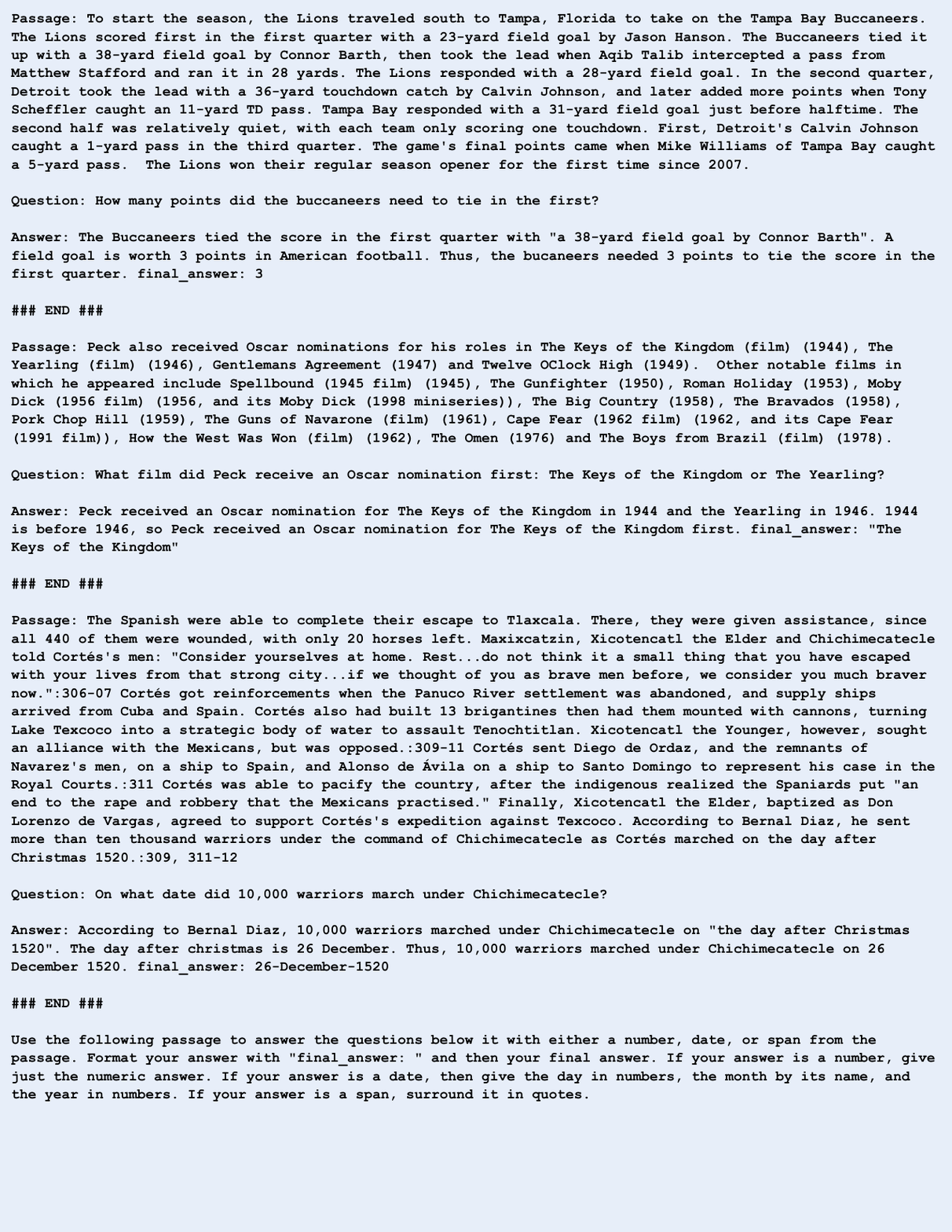}
\caption{Initialization prompt for Question Answering}
\end{figure*}

\end{document}